\documentclass[10pt,twocolumn,letterpaper]{article}

\usepackage{cvpr}              

\usepackage{graphicx}
\usepackage{amsmath}
\usepackage{amssymb}
\usepackage{booktabs}
\usepackage{bm}
\usepackage{multirow}
\usepackage{multicol}
\usepackage{color}
\usepackage[accsupp]{axessibility}
\usepackage[pagebackref,breaklinks,colorlinks,linkcolor=red]{hyperref}

\definecolor{cGreen}{RGB}{47,139,69}
\definecolor{Red}{RGB}{255,0,0}
\def\corres{\footnote{Corresponding author}}

\usepackage[capitalize]{cleveref}
\crefname{section}{Sec.}{Secs.}
\Crefname{section}{Section}{Sections}
\Crefname{table}{Table}{Tables}
\crefname{table}{Tab.}{Tabs.}

\def\cvprPaperID{9651} 
\def\confName{CVPR}
\def\confYear{2022}

\begin{document}

\title{Transformer Tracking with Cyclic Shifting Window Attention}

\author{Zikai Song\textsuperscript{1}\qquad Junqing Yu\textsuperscript{1}\corres{}\qquad Yi-Ping Phoebe Chen\textsuperscript{2}\qquad Wei Yang\textsuperscript{1}\\
\textsuperscript{1}Huazhong University of Science and Technology, China\qquad\textsuperscript{2}La Trobe University, Australia\\
{\tt\small \{skyesong, yjqing, weiyangcs\}@hust.edu.cn, phoebe.chen@latrobe.edu.au}
}
\maketitle

\renewcommand{\thefootnote}{}
\footnotetext{*~Corresponding author}

\begin{abstract}
Transformer architecture has been showing its great strength in visual object tracking, for its effective attention mechanism. Existing transformer-based approaches adopt the pixel-to-pixel attention strategy on flattened image features and unavoidably ignore the integrity of objects. In this paper, we propose a new transformer architecture with multi-scale cyclic shifting window attention for visual object tracking, elevating the attention from pixel to window level. The cross-window multi-scale attention has the advantage of aggregating attention at different scales and generates the best fine-scale match for the target object. Furthermore, the cyclic shifting strategy brings greater accuracy by expanding the window samples with positional information, and at the same time saves huge amounts of computational power by removing redundant calculations. Extensive experiments demonstrate the superior performance of our method, which also sets the new state-of-the-art records on five challenging datasets, along with the VOT2020, UAV123, LaSOT, TrackingNet, and GOT-10k benchmarks. Our project is available at \url{https://github.com/SkyeSong38/CSWinTT}.
 
\end{abstract}

\section{Introduction}
\label{sec:intro}

\begin{figure}
  \centering
  \begin{subfigure}{0.675\linewidth}
    \includegraphics[width=\linewidth,keepaspectratio,page=5]{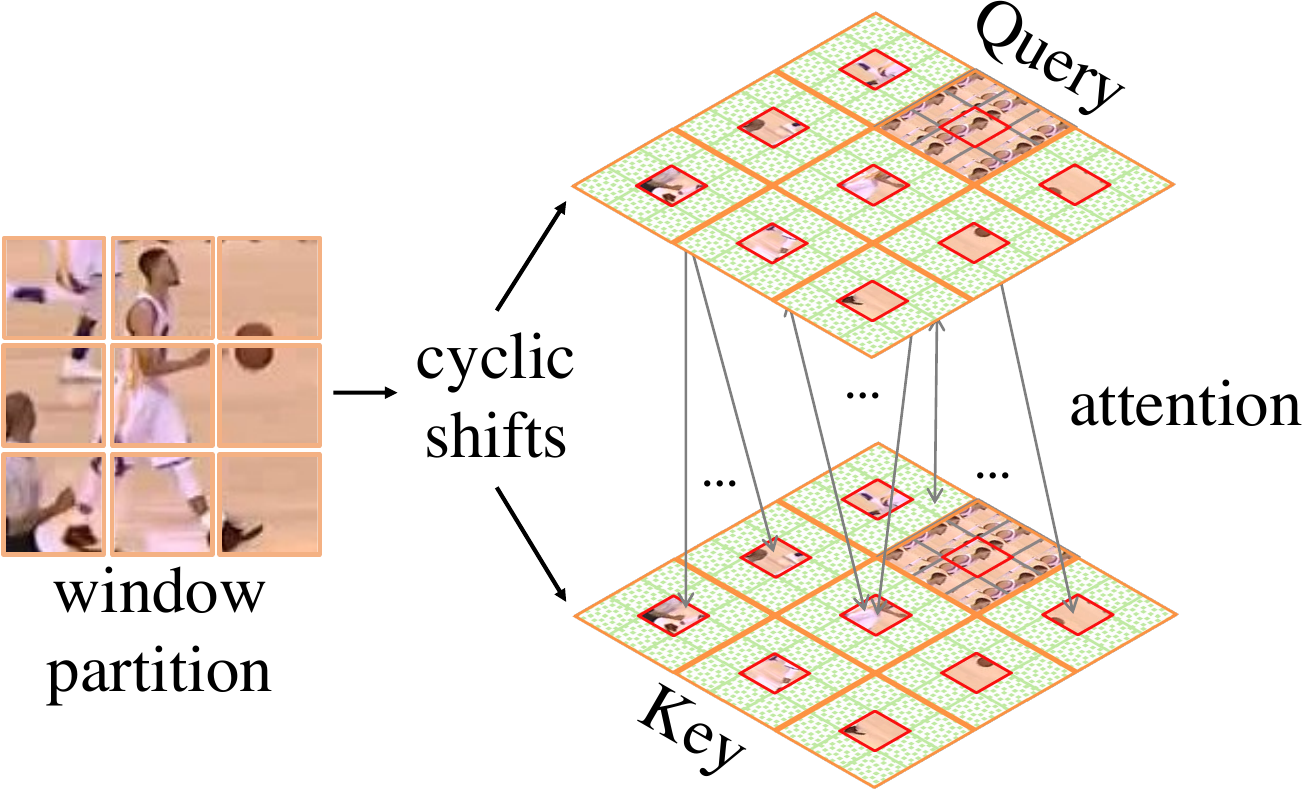}
    \caption{Cyclic shifting window-level Attn}
    \label{fig:1a}
  \end{subfigure}
  \hfill
  \begin{subfigure}{0.315\linewidth}
    \includegraphics[width=\linewidth,keepaspectratio,page=4]{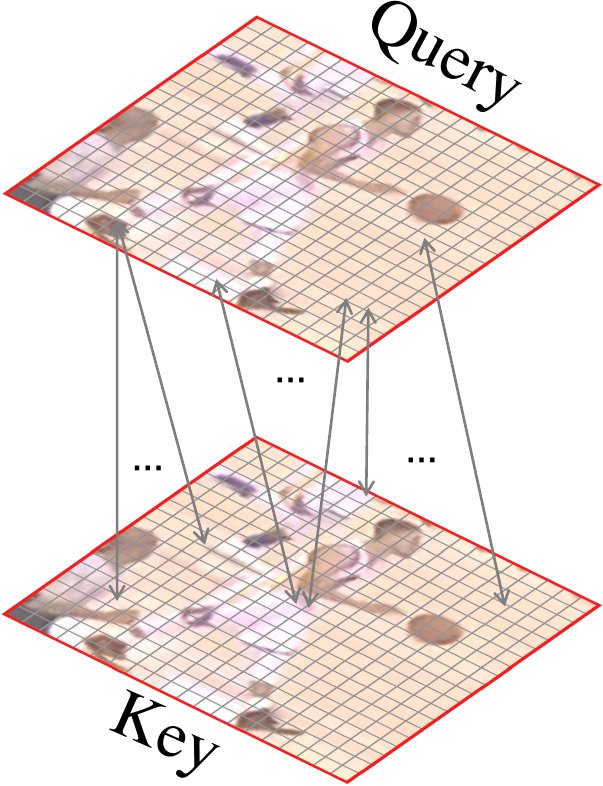}
    \caption{Pixel-level Attn}
    \label{fig:1b}
  \end{subfigure}
  \caption{(a) The proposed approach firstly achieves the window-level attention between query and key through window partitioning, and then applies cyclic shifts for each window (from the base sample in the red box to generated samples surrounded by orange boxes) to greatly extend the number of window samples, while maintaining the integrity of objects. (b) Previous transformers produce pixel-level attention, which weakens the positional information between pixels and ignores the integrity of objects.}
  \label{fig:1}
\end{figure}

Visual object tracking (VOT) is one of the fundamental problems in computer vision research with a wide range of applications in video surveillance, autonomous vehicles, human-machine interaction, and others. It aims to estimate the position of a target object in each video frame, commonly represented as a bounding box encapsulating the target. The target object is given as a template in the initial frame, and the tracker is required to extract proper features about the target and localize the target in the following frames. Most of the popular trackers~\cite{SiamFC,SiamRPN, SiamMask, SiamRPN++,SiamRCNN} adopt the Siamese network structure, which conducts tracking by calculating the similarity between the template and search region in the current frame. The similarity metric of cross-correlation used in Siamese trackers is prone to lose much semantic information for it is a single-level linear computational process. This deficiency can be well tackled by using the attention mechanism to learn the global context. Recently, transformer-based approaches~\cite{ViT,SwinTransformer,DERT,Max-deeplab} have reported new state-of-the-art performance on image recognition, object detection, and semantic segmentation benchmarks. This is no wonder as transformer\cite{Transformer} has a powerful cross-attention mechanism to reasoning between patches\cite{COTR}. Particularly, transformer trackers~\cite{TransT,TMT,TrTr,STARK} have shown their great strength by introducing the attention mechanism to enhance and fuse the features of the target and the tracked object. However, we observe that these transformer trackers simply put the flattened features of the template and search region into pixel-level attention, each pixel of a flattened feature (Query) matches all pixels of another flattened feature (Key) in a complete and disordered manner, as shown in Figure \ref{fig:1b}. This pixel-level attention destroys the integrity of the target object and leads to information loss of relative positions between pixels.

In this paper, we propose a novel multi-scale cyclic shifting window transformer for visual object tracking to further lift pixel-level attention to window-level attention, calculating attention between indivisible windows by treating each window as a whole keeps the location information within the window. The proposed method is inspired by the seminal work of the Swin Transformer\cite{SwinTransformer}, which adopts a hierarchical transformer structure by starting from small-sized patches and gradually increasing the size through merging to achieve a broader receptive field. Different from the Swin Transformer, we calculate the cross-window attention between the template and search region directly, which helps to discriminate the target from the background by ensuring the integrity of the object. Further, we propose a multi-head multi-scale attention where each head of the transformer measures the relevance among partitioned windows at a specific scale. The key idea here is to apply a cyclic shifting strategy on each window, as shown in Figure~\ref{fig:1a}, for generating more accurate attention results. To address performance drop around boundaries caused by the cyclic shifting operation, we design a spatially regularized attention mask which turns out to be very effective in alleviating the boundary artifacts. Finally, we present some efficient computation strategies to avoid redundant computation introduced by multi-scale cyclic shifting windows, which greatly reduce the time and computational cost. Extensive experiments demonstrate that our tracker performs remarkably better than other state-of-the-art algorithms.

To summarize, our main contributions include:
\begin{enumerate}
\item We propose a novel transformer architecture with multi-scale cyclic shifting window attention for visual object tracking, uplifting the original pixel-level attention to the new deliberately designed window-level attention. The cross-window attention ensures the integrity of the tracking object, and the cyclic shifts bring greater accuracy by expanding window samples.

\item We design a spatially regularized attention mask and some computational optimization strategies to improve the accuracy and speed of the window attention. Specifically, a spatially regularized attention mask is used to address performance drop around boundaries caused by the cyclic shifts, and we propose three computational optimization strategies to remove redundant computations.

\end{enumerate}

\section{Related Work}

\label{sec:related}
\noindent \textbf{Visual object tracking.}
Existing visual object tracking approaches can be roughly divided into two categories, the Correlation Filter (CF) based trackers and Deep Neural Network (DNN) based trackers. CF based approaches~\cite{MOSSE,KCF,BACF,LADCF,CCOT,ECO,UPDT} exploit the convolution theorem and train a filter in the Fourier domain that maps known target images to the desired output. The filter is learned through circular shifting patches around the target object to discriminate background against the target. DNN based trackers refer to the methods adopting deep neural networks in the tracking process. Many methods~\cite{MDNet, RTMDNet, DATRL} treat the tracking task as a basic recognition task, i.e., using a convolutional backbone network to extract features and locate the target by classification heads in the form of fully connected layers.

Recent years, tracking algorithms that adopt a Siamese network structure \cite{SiamFC,ATOM,DiMP,PrDiMP,SiamRPN,DaSiamRPN,SiamMask,SiamRPN++,RPT,SiamAttn, SiamDW, siamfc++,SiamGAT} have shown great success. A Siamese network usually consists of two branches, one for template and the other for search regions, and similarities between them are reported through cross-correlations. However, such a strategy is unable to effectively explore the semantic correlation between template and search regions. This issue leads to the further exploration of using the powerful cross-attention mechanism of transformer structure for object tracking.

\label{sec:method}
\begin{figure*}
  \centering
  \includegraphics[width=\linewidth,keepaspectratio,page=1]{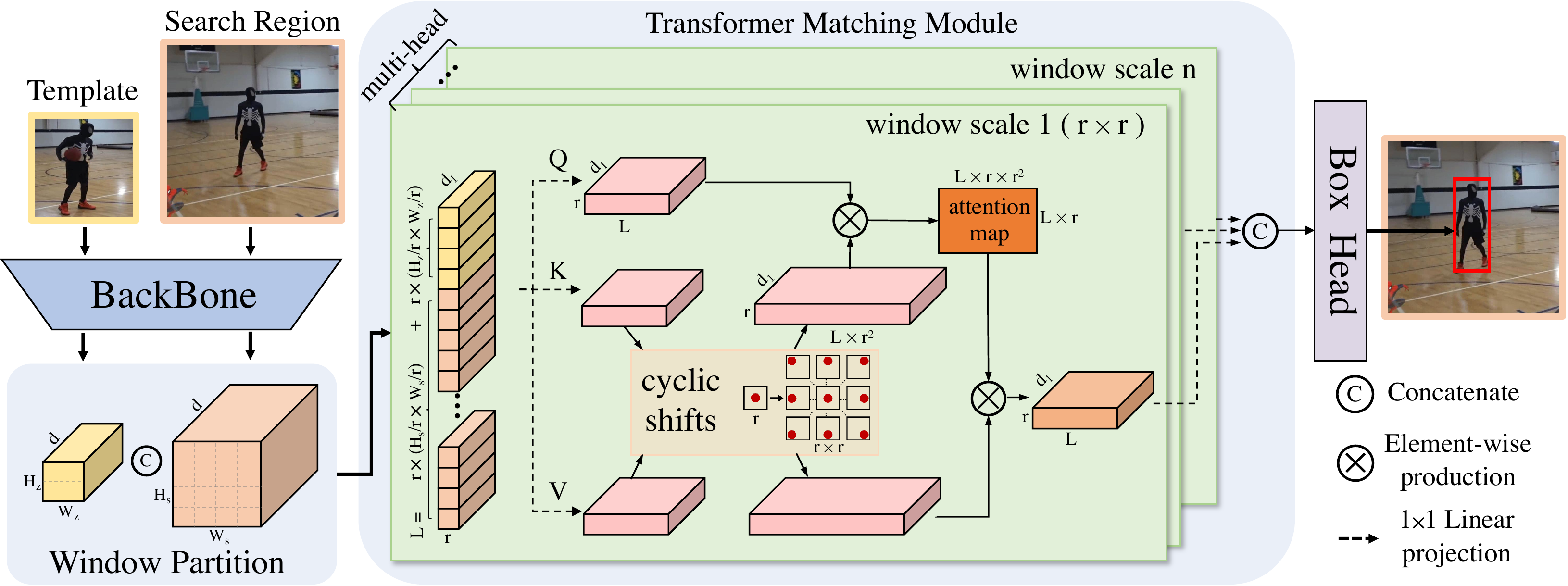}
  \caption{An overview of proposed tracking architecture. Given a template image in the first frame and a search region in the subsequent frame, we extract deep feature maps through a backbone network. Then these two features are partitioned into small windows and flattened as a window sequence. The transformer matching module applies window-level attention to the window sequence. Specifically, the Key(K)-Value(V) pair in the transformer employ the cyclic shifting strategy to generate a great number of samples. Then the transformer outputs the fused features that contain the deep matching information between template and search region, these fused features are passed through the bounding box prediction head to obtain the final tracking result.}
  \label{fig:2}
\end{figure*}

\noindent \textbf{Vision transformers.}
Vaswani $et~al$.~\cite{Transformer} propose the very first Transformer structure for handling long-range dependencies in Natural Language Processing (NLP). The basic block in a transformer is the attention module, which takes a sequence as input and measures the relevance of different parts of the sequence, aggregating the global information from the input sequence. Transformer not only conducts the self-attention within a single input but also calculates the cross-attention between different inputs. ViT\cite{ViT} first introduces transformer to image recognition tasks. Ever since, transformer has been widely applied in image classification\cite{ViT,SwinTransformer}, object detection\cite{DERT}, semantic segmentation\cite{Max-deeplab}, visual object tracking\cite{STARK,TransT,TransCenter,TMT} and etc.
 
The seminal work of Swin Transformer\cite{SwinTransformer} proposes an effective hierarchical architecture with shifted windows and achieves the state-of-the-art performance on COCO object detection\cite{COCO}, and ADE20K semantic segmentation\cite{ADE20K}. Though our approach is inspired by the Swin Transformer, we have three fundamental differences: (1) where attention is applied are different. Swin Transformer partitions the image into windows and then conducts pixel-level attention inside each window, while we do window partitioning in feature maps, and calculate attention between windows by treating each window as a whole. (2) multi-scaling strategy is different. Swin transformer uses the same window size in one layer and merges windows to form a larger window in deeper layers. In contrast, we use windows with different sizes as heads for multi-scale matching. (3) window shifting is applied differently. Swin Transformer shifts the whole feature map, in order to exchange information and provide connectivity between different windows. We apply independent cyclic shifts in each window in a non-exchangeable way. Additionally, in contrast to Swin Transformer, where each window is shifted only once, in our algorithm each window is shifted multiple times depending on its size.

Recently, transformer-based visual object tracking methods have become more and more popular. TrDiMP~\cite{TMT} separates the encoder-decoder transformer into two Siamese-like branches, the encoder reinforces the template features and the decoder propagates the tracking cues from previous templates to the current frame. TransT\cite{TransT} proposes a feature fusion network and employs an attention mechanism to combine the features of template and search region. This feature fusion network consists of an ego-context augment module based on self-attention and a cross-feature augment module based on cross-attention. STARK\cite{STARK} develops a spatial-temporal architecture based on the encoder-decoder transformer, the encoder learns the relationship between template and search region and the decoder learns a query embedding to predict the target positions. Moreover, STARK introduces a corner-based prediction head used for estimating the bounding box and a score head for controlling the updates of the template image. Most of the previous tracking algorithms such as \cite{TMT,STARK} use encoder-decoder structure to enhance or fuse the features, while we consider the transformer as a feature matching module to calculate the similarity between template and search region. Moreover, previous approaches use the transformer naively and do touch the attention mechanism within. On the contrary, we carefully design a multi-head multi-scale window-level attention transformer with the cyclic shifting strategy, to fully exploit the transformer structure for object tracking.

\section{Method} 
In this section, we present our multi-scale \textbf{c}yclic \textbf{s}hifting \textbf{win}dow \textbf{t}ransformer \textbf{t}racker, namely CSWinTT. We use the transformer as a matching module for measuring the relevance between template and search region to fully exploit the powerful cross-attention capabilities of the transformer.

The tracking architecture is visualized in Figure \ref{fig:2}, which consists of three major components: a feature extraction backbone, a transformer matching module, and a bounding box estimation head. We choose the ResNet-50\cite{ResNet} as our backbone for feature extraction, which takes a pair of image inputs, i.e., the template image and the search region image. The pair of output features are then partitioned into window sequences and fed into the transformer matching module. The matching module concatenates the two window sequences and sends them to the multi-head 6-layer transformer. The multi-head transformer uses a specific window size for each head for scale adaption. Finally, the outputs of each transformer head are concatenated together, passed through the corner-based box estimation head as in~\cite{STARK} to get the result bounding box.

\subsection{Multi-Scale Cyclic Shifting Window Attention}
%
\noindent \textbf{Multi-scale window partition.}
Give template patch $z$ in the initial frame and an image $s$ for search region. We pass $z$ and $s$ through the backbone and a bottleneck layer, obtaining the feature map $f_z\in\mathbb{R}^{d\times H_z \times W_z}$ and $f_s\in\mathbb{R}^{d\times H_s \times W_s}$ respectively. Then we extract patches of shape $r_i \times r_i$ from $f_z$ and $f_s$ in head $i$, where $d_i$ represents the number of channels of head $i$. The total number of template windows is $N_z^i = \frac{H_z}{r_i} \times \frac{W_z}{r_i}$ and for search region windows is $N_s^i = \frac{H_s}{r_i} \times \frac{W_s}{r_i}$. After window extraction, the feature maps are reshaped to window sequences $f_z^i \in \mathbb{R}^{N_z^i \times d_i \times r_i \times r_i}$ and $f_s^i \in \mathbb{R}^{N_s^i \times d_i \times r_i \times r_i}$. The two window sequences are then concatenated along the spatial dimension and generate $f_c^i$ with $(N_z^i + N_s^i) \times d^i$. Then query-key-value attention mechanism is applied with query $\mathbf{Q_i}$, key $\mathbf{K}_i$ and value $\mathbf{V}_i$. Key-value pairwise similarities are then calculated and fused using a multi-head attention mechanism as follows:

\noindent \textbf{Multi-head attention.} Multi-head attention is the fundamental component in our architecture. As described in \cite{Transformer}, given queries $\mathbf{Q}$, keys $\mathbf{K}$, and values $\mathbf{V}$. The multi-head attention is computed as:

\begin{equation}
\begin{split}
  \text{MultiHead}(\mathbf{Q}, \mathbf{K}, \mathbf{V}) &= \text{Concat}(\mathbf{H}_1,\ldots, \mathbf{H}_{n_h}) \\
  \text{where} \quad \mathbf{H}_i&=\text{Attention}(\mathbf{Q}_i, \mathbf{K}_i, \mathbf{V}_i) \\
  &= \text{softmax}(\frac{\mathbf{Q}_i \mathbf{K}_i^T}{\sqrt{d_k}})\mathbf{V}_i
\end{split}
\label{eq:1}
\end{equation}
where $n_h$ is the number of heads, and $d_k$ is the dimension of key. For a clearer description of the post-order steps, we define the attention score as:
\begin{equation}
  \text{AttnScore}(\mathbf{Q}, \mathbf{K})= \frac{\mathbf{Q}\mathbf{K}^T}{\sqrt{d_k}}
  \label{eq:2}
\end{equation}

\noindent \textbf{Cyclic shifting strategy.}
Compared to the pixel-level attention, one problem of our window attention is the resolution of attention map reduces from $\mathbb{R}^{(H_z W_z + H_s W_s)^2}$ to $\mathbb{R}^{(N_z^i + N_s^i)^2}$. This will lead to a coarser similarity score, and it is hard to fuse the output of each head for the attention generated by different heads does not have the same size.

\begin{figure}
    \centering
    \includegraphics[width=\linewidth,keepaspectratio,page=3]{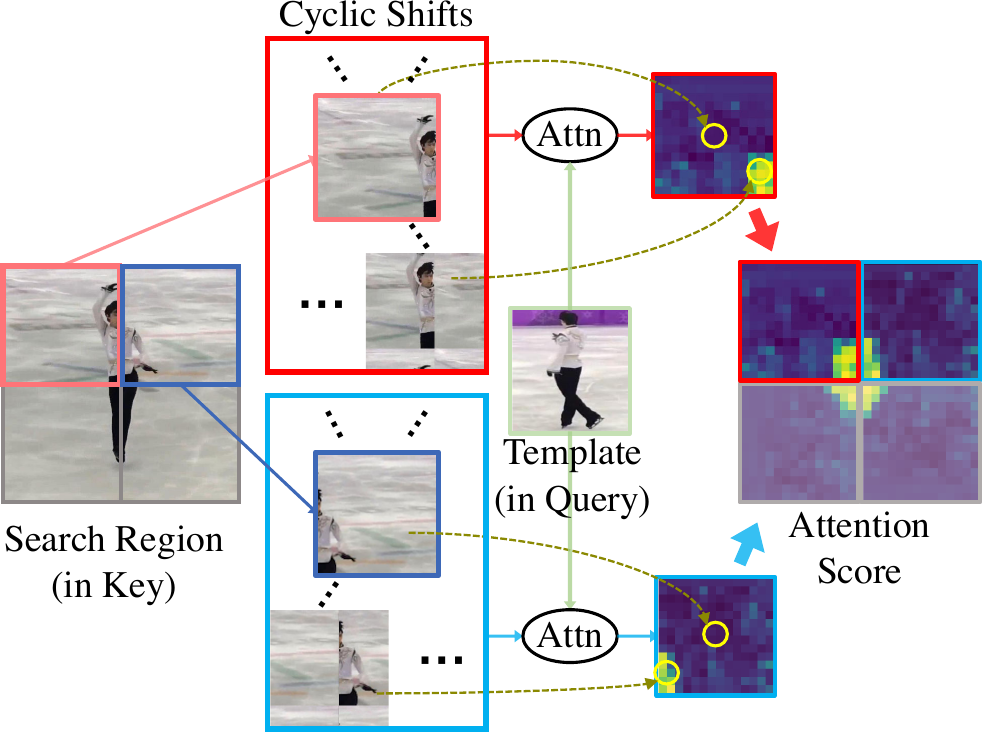}
    \caption{Cyclic shifting window attention. We illustrate that when the search region in the Key is partitioned into 4 windows and the template in the Query becomes one window. The above two windows in the search region are selected as examples to do attention with the template after cyclic shifts, and attention scores are obtained respectively, then they are combined to form the final attention score.}
    \label{fig:cs}
\end{figure}

To address the aforementioned problems, we propose a cyclic shifting strategy on the proposed window-level attention. It enhances the effectiveness of cross-window attention while preserving positional information and the integrity of objects, as shown in Figure \ref{fig:cs}. Within a specific head, consider a window with size $r\times r$ referred to as the base sample. We define the shift operator $\text{shift}(x,y)$ of the base sample as translating the sample by $x$ pixels in the horizontal direction, and $y$ pixels in the vertical direction. Our cyclic shifts of a base sample then are performed at a single-pixel distance and move the sample into bottom-right directions with boundaries being warped back to the top-left. The operation $\text{shift}(x,y), x,y \in [-r+1,r-1]$ generates $r \times r$ into $(2r-1)^2$ samples for the base sample with window size $r \times r$. Obviously, these cyclic shifts generate a lot of duplicates, we will discuss how to effectively remove duplicate computations in the section \ref{sec:comopt}.

\subsection{Efficient Computation} 
\begin{figure}
    \centering
    \includegraphics[width=0.9\linewidth,keepaspectratio,page=2]{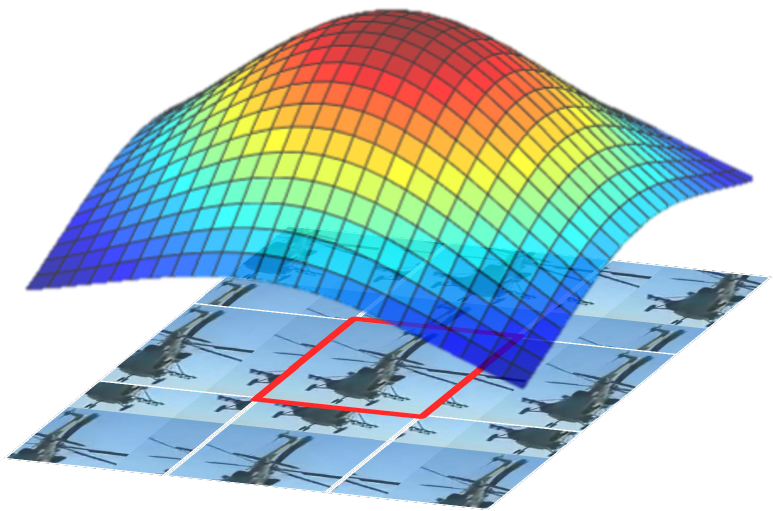}
    \caption{Visualization of the spatial regularization weights designed to alleviate the boundary artifacts. The red box in the center is the base sample, generated samples are penalized more if they are further from the center base sample (i.e., the smaller the value of their applied attention mask).}
    \label{fig:4}
\end{figure}

\noindent \textbf{Spatially regularized attention mask.} 
In practice, we find the shifted samples near the center contribute more to the final attention. This is reasonable as samples close to the boundaries are more likely to break the integrity and position information of the tracking object in the window. Hence we design a weighting scheme applied as a form of attention mask $\mathbf{M}$ in the transformer, as shown in Figure \ref{fig:4}. The spatial weights of the mask penalize samples depending on their spatial locations, the formula for weight generation is expressed in \ref{eq:sr}. The further away the generated sample is from the base sample, the larger the penalty is and the weight is smaller. 

\begin{equation}
  \mathbf{M}(x,y) = -(\frac{x}{r})^2-(\frac{y}{r})^2,\quad x,y\in[-r+1,r-1]
  \label{eq:sr}
\end{equation}
\noindent The spatial attention mask $\mathbf{M}$ is directly added onto the attention score in \ref{eq:2}.

Recall that the sizes of query $\mathbf{Q}_i$ and key $\mathbf{K}_i$ in the $i$-th head are $\mathbb{R}^{(N_z+N_s) \times d_i \times r_i \times r_i}$ respectively. With $r_i \times r_i$ as the window size, and each window is expanded into $[(2r_i-1)^2 \times d_i \times  r_i \times r_i]$ with cyclic shifts. The size of attention score on each window is $[(2r_i-1)^2 \times (2r_i-1)^2]$, this is formed by the window of query and key. Since the query and key are also cyclic shifting, we add the spatial weights of size $[(2r_i-1) \times (2r_i-1)]$ to the last dimension of attention score (dimension of keys), achieving the effect of positional penalty.

\noindent \textbf{Computational optimization.} 
\label{sec:comopt}
Intuitively, the cyclic shifts increase the computational cost greatly, especially when the window size is large. To achieve computational efficiency, we optimize in three ways: (i) eliminating the cyclic shifts of the Query; (ii) halving the duplicated shifting periods; and (iii) adopting the programming optimization for matrix translation.

Suppose we have $\mathbf{Q}$, $\mathbf{K}$ and $\mathbf{V}$ of size $(H, W, d)$. Standard transformer flattens the features to $(HW, d)$, two parts account for the time cost of attention computation are the attention score computation $O(HW \times d HW)$ and fusion feature computation $O(HW\times HW \times d)$. After applying the cyclic shifts, the size of $\mathbf{Q}$, $\mathbf{K}$ and $\mathbf{V}$ are $(\frac{H}{r},\frac{H}{r},2r-1,2r-1,r,r,d)$ with $r$ as window size, and the complexity of computing the attention score increases to $O((\frac{H}{r} \frac{W}{r} (2r-1) (2r-1))^2\times r^2d)$. We observe if $Q$ and $K$ perform the same shifting, computing attention scores is meaningless, so we just need to perform a cyclic operation on $K$ and keep $Q$ unchanged to achieve the same effect. In addition, note that the cyclic generated samples to the bottom-right and the top-left directions are repeated, we reduce the number of shifting periods by half for better efficiency. And we also apply a programming trick to improve the tracking speed, which is using permutations of matrix coordinates to perform cyclic shifts instead of direct translations on the matrix.

\subsection{Tracking with Window Transformer}

Multi-scale window transformer facilitates the tracking process by conducting accuracy-aware attention with windows at different scales. Therefore, the choice of the window sizes is extremely important. In our implementation, we set the number of heads $n_h$ to $8$ with window size $r_i = [1,2,4,8,1,2,4,8]$ for head $i$. Notice the second half of the heads have the same window size, that's because we adopt feature map translation which displaces the backbone feature of the search image by $(\frac{r_i}{2},\frac{r_i}{2})$ pixels. In this way, when the windows are partitioned in a non-overlapping manner, the contents of windows are complemented by each other to avoid the situation that the object has been segmented all the time.

In further, to improve the robustness of the tracking algorithm, we use two templates of the same size as the input of the transformer. One of which is fixed using the initial template, the other is online updated to the latest tracking result with high confidence, a score head is employed to control the updates, as designed in STARK\cite{STARK}.

In the training stage, we use the L1 loss and the generalized IoU loss \cite{giou} to train the overall architecture in an end-to-end manner. During the inference, the template image and its corresponding backbone features are initialized in the first frame, and the search region is used as the input to the tracker during the tracking process in subsequent frames, with the predicted bounding box returned by the network as the final result.

\begin{table*}[htbp]\small
  \centering
  \caption{Comparisons on four tracking benchmarks. The \textcolor{red}{\textbf{red}}, \textcolor{cGreen}{green} and \textcolor{blue}{blue} indicate performances ranked at first, second, and third places}
  \label{tab:1}
  \setlength{\tabcolsep}{2mm}{  
  \begin{tabular}{lc cc c ccc c ccc c ccc}
    \toprule
    \multirow{2}*{Method} & \multirow{2}*{Year} & \multicolumn{2}{c}{UAV123\cite{UAV123}}&& \multicolumn{3}{c}{LaSOT\cite{LaSOT}} && \multicolumn{3}{c}{TrackingNet\cite{TrackingNet}} && \multicolumn{3}{c}{GOT-10k\cite{GOT-10k}}\\
    \cline{3-4}
    \cline{6-8}
    \cline{10-12}
    \cline{14-16}
    & & AUC&P && AUC&P$_{Norm}$&P && AUC&P$_{Norm}$&P && AO&SR$_{0.5}$&SR$_{0.75}$\\
    \midrule[0.5pt]
    SiamFC\cite{SiamFC}&2016& 49.2&72.7 && 33.6&42.0&33.9 && 57.1&66.3&53.3 && 34.8&35.3&9.8 \\
    ECO\cite{ECO}&2017& 52.5&74.1 && 32.4&33.8&30.1 && 55.4&61.8&49.2 && 31.6&30.9&11.1 \\
    ATOM\cite{ATOM}&2019& 61.7&82.7 && 51.5&57.6&50.5 && 70.3&77.1&64.8 && 55.6&63.4&40.2 \\
    DiMP\cite{DiMP}&2019& 64.2&84.9 && 57.7&66.4&57.9 && 74.0&80.1&68.7 && 61.1&71.7&49.2 \\
    SiamRPN++\cite{SiamRPN++}&2019& 64.2&84.0 && 49.6&56.9&49.1 && 73.3&80.0&69.4 && 51.7&61.6&32.5 \\
    SiamFC++\cite{siamfc++}&2020& 61.8&80.4 && 54.4&62.3&54.7 && 75.4&80.0&70.5 && 59.5&69.5&47.9 \\
    D3S\cite{D3S}&2020& -&- && -&-&- && 72.8&76.8&66.4 && 59.7&67.6&46.2 \\
    MAML\cite{MAML}&2020& -&- && 52.3&-&53.1 && 75.7&82.2&72.5 && -&-&- \\
    SiamAttn\cite{SiamAttn}&2020& 65.0&84.5 && 56.0&64.8&- && 75.2&81.7&- && -&-&- \\
    KYS\cite{KYS}&2020& -&- && 55.4&63.3&55.8 && 74.0&80.0&68.8 && 63.6&75.1&51.5 \\
    PrDiMP\cite{PrDiMP}&2020& 66.6&87.2 && 59.9&68.8&60.8 && 75.8&81.6&70.4 && 63.4&73.8&54.3 \\
    Ocean\cite{OCEAN}&2020& 62.1&82.3  && 51.6&60.7&52.6 && 69.2&79.4&68.7 && 61.1&72.1&47.3 \\
    SiamRCNN\cite{SiamRCNN}&2020& -&- && 64.8&72.2&- && 81.2&85.4&\textcolor{cGreen}{80.0} && 64.9&72.8&59.7 \\
    SiamGAT\cite{SiamGAT}&2021& 64.6&84.3 && 53.9&63.3&53.0 && -&-&- && 62.7&74.3&48.8  \\
    AutoMatch\cite{AutoMatch}&2021& 64.4&83.8 && 58.2&67.5& 59.9 && 76.0&82.4&72.5 && 65.2&76.6&54.3 \\
    TrDiMP\cite{TMT}&2021& 67.0&\textcolor{blue}{87.6} && 64.0&73.2&66.6 && 78.4&83.3&73.1 && \textcolor{cGreen}{68.8}&\textcolor{red}{\textbf{80.5}}&59.7 \\
    TransT\cite{TransT}&2021& \textcolor{blue}{68.1}&\textcolor{blue}{87.6} && \textcolor{blue}{64.9}&\textcolor{blue}{73.8}&\textcolor{blue}{69.0} && \textcolor{cGreen}{81.4}&\textcolor{red}{\textbf{86.7}}&\textcolor{red}{\textbf{80.3}} && 67.1&76.8&\textcolor{blue}{60.9} \\
    STARK-ST50\cite{STARK}&2021& \textcolor{cGreen}{69.2}&\textcolor{cGreen}{88.2} && \textcolor{cGreen}{66.0}&\textcolor{red}{\textbf{75.5}}&\textcolor{cGreen}{70.8} && \textcolor{blue}{81.3}&\textcolor{blue}{86.1}&- && \textcolor{blue}{68.0}&\textcolor{blue}{77.7}&\textcolor{cGreen}{62.3}  \\
    \midrule[0.1pt]
    CSWinTT&Ours& \textcolor{red}{\textbf{70.5}}&\textcolor{red}{\textbf{90.3}} && \textcolor{red}{\textbf{66.2}}&\textcolor{cGreen}{75.2}&\textcolor{red}{\textbf{70.9}} && \textcolor{red}{\textbf{81.9}}&\textcolor{red}{\textbf{86.7}}&\textcolor{blue}{79.5} && \textcolor{red}{\textbf{69.4}}&\textcolor{cGreen}{78.9}&\textcolor{red}{\textbf{65.4}} \\
  \bottomrule
\end{tabular}
}
\end{table*}

\section{Experiments}

\subsection{Implementation Details}
We train our model on the LaSOT\cite{LaSOT}, GOT-10k\cite{GOT-10k}, and TrackingNet\cite{TrackingNet} datasets. The image pairs are directly sampled from the same sequence and common data augmentation operations including brightness jitter and horizontal flip are applied. The size of the input template is 128$\times $128 pixels, the search region is $5^2$ times of the target box area and further resized to 384$\times $ 384 pixels. We use ResNet-50\cite{ResNet} as the backbone, the parameters of which are initialized with ImageNet pretrained\cite{imagenet} model. Other parameters in our model are initialized with Xavier Uniform. We use the $\lambda_{l1}=5$ and $\lambda_{giou}=2$ as the loss weight for l1 loss and giou loss\cite{giou}.  The AdamW optimizer \cite{adamw} is employed with initial learning rates of 1e-5 and 1e-4 for backbone parameters and other parameters, respectively, and weight decay is set to 1e-4 for every 10 epochs after 500 epochs. We train our model on two Nvidia Tesla T4 GPUs for a total of 600 epochs, each epoch uses $4 \times 10^4$ images. The mini-batch size is set to 64 images with each GPU hosting 32 images. The training process of the update module is the same as \cite{STARK}. Our approach is implemented in Python 3.7 using PyTorch 1.6. CSWinTT operates about 12 frames per second (FPS) on a single GPU during the online tracking process.

\subsection{State-of-the-art Comparison}
We compare our proposed CSWinTT algorithm with the state-of-the-art trackers on five tracking benchmarks, including UAV123\cite{UAV123}, LaSOT\cite{LaSOT}, TrackingNet\cite{TrackingNet}, GOT-10k\cite{GOT-10k}, and VOT2020\cite{VOT2020}. 

\begin{table}[htbp]\normalsize
  \centering
  \caption{Result comparisons on VOT2020\cite{VOT2020}, where trackers only predict bounding boxes rather than reporting masks.}
  \label{tab:2}
  \setlength{\tabcolsep}{2mm}{
    \begin{tabular}{lccc}
    \toprule
    &EAO$\uparrow$&Accuracy$\uparrow$&Robustness$\uparrow$\\
    \midrule[0.5pt]
    KCF\cite{KCF}&0.154&0.407&0.430 \\
    SiamFC\cite{SiamFC}&0.179&0.418&0.502 \\
    CSR-DCF\cite{CSR-DCF}&0.193&0.406&0.582 \\
    ATOM\cite{ATOM}&0.271&0.462&0.734 \\
    DiMP\cite{DiMP}&0.274&0.457&0.740 \\
    UPDT\cite{UPDT}&0.278&0.465&0.755 \\
    TrDiMP\cite{TMT}& \textcolor{blue}{0.300}&0.471&\textcolor{cGreen}{0.782} \\
    TransT\cite{TransT}& 0.293&\textcolor{blue}{0.477}&0.754 \\
    STARK\cite{STARK}& \textcolor{cGreen}{0.303}&\textcolor{red}{\textbf{0.481}}&\textcolor{blue}{0.775} \\
    \midrule[0.1pt]
    CSWinTT(Ours)& \textcolor{red}{\textbf{0.304}}&\textcolor{cGreen}{0.480}&\textcolor{red}{\textbf{0.787}} \\
    \bottomrule
    \end{tabular}
  }
\end{table}

\noindent \textbf{UAV123}\cite{UAV123}: UAV123 gathers an application-specific collection of 123 sequences and captures from unmanned aerial vehicles video dataset. It adopts the Area Under the Curve (AUC) and Precision (P) as the evaluation metrics. The precision is used to measure the center distance and the AUC plot computes the intersection-over-union (IoU) score between the estimated bounding box and the ground-truth. As shown in Table \ref{tab:1}, where the previous state-of-the-art trackers such as TrDiMP\cite{TMT}, TransT\cite{TransT}, and START\cite{STARK} are included for comparison, note that STARK-ST50 is chosen for the reason that it uses the same ResNet-50 backbone as our algorithm, which can more fairly compare the performance of transformer structure. Our CSWinTT outperform the aforementioned methods by a considerable margin and exhibits very competitive performance (70.5\% AUC and 90.3\% Precision) when compared to the best previous tracker STARK (69.2\% AUC and 88.2\% Precision)

\noindent \textbf{LaSOT}\cite{LaSOT}: LaSOT is a large-scale long-term dataset including 1400 sequences and distributed over 14 attributes, the testing subset of LaSOT contains 280 sequences with an average length of 2448 frames. Methods are ranked by the AUC, Precision, and Normalized Precision (P$_{Norm}$). The evaluation results of compared tracking algorithms are shown in Table \ref{tab:1}. Our model achieves the top-rank AUC score (66.2\%) and Precision score (70.9\%), which outperforms the previous best result by STARK-ST50, and also surpasses the other two transformer trackers TransT\cite{TransT}/TrDiMP\cite{TMT} for 1.3\%/2.2\% AUC score, respectively.

\noindent \textbf{TrackingNet}\cite{TrackingNet}: TrackingNet is a large-scale tracking dataset consisting of 511 sequences for testing. The evaluation is performed on the online server. \ref{tab:1} shows that, compared with SOTA models, our CSWinTT performs better visual tracking quality and ranks at the first in AUC score of 81.9\% and normalized precision of 86.7\%. The specific gain is 0.7\% relative improvement of the AUC score when compared with the TransT\cite{TransT}, which represents the previous best algorithm on this benchmark.

\noindent \textbf{GOT-10k}\cite{GOT-10k}: GOT-10k is a large-scale dataset containing over 10k videos for training and 180 for testing. It forbid the trackers to use external datasets for training. We follow this protocol by retraining our trackers using only the GOT10k train split. As can be seen from Table \ref{tab:1}, among previous transformer trackers, TrDiMP~\cite{TMT} and STARK-ST50 \cite{STARK} provides the best performance, with an AUC score of 68.8\% and 68.0\%. Our approach has remarked improvement and obtains an AUC score of 69.4\%, significantly outperforming the best existing tracker (TrDiMP) by 0.6\%.

\noindent \textbf{VOT2020}\cite{VOT2020}: VOT2020 benchmark contains 60 challenging videos. The performance on this dataset is evaluated using the expected average overlap (EAO), which takes both accuracy (A) and robustness (R) into account. In addition, a new anchor-based evaluation protocol is proposed in VOT2020, the segmentation mask is adopted as the ground-truth. However, since our algorithm does not output a segmentation mask, trackers only predict bounding boxes are chosen as the comparisons to ensure a fair evaluation. It can be seen from the data in Table \ref{tab:2} that CSWinTT obtains an EAO of 0.304, ranking first in previous trackers.

\subsection{Ablation Study}
We conduct ablation analysis to evaluate the different components in our CSWinTT and evaluate the performance of diverse window sizes using the UAV123 dataset\cite{UAV123}. Besides, we show the superiority of the three previously mentioned computational optimization strategies.

\begin{table}[htbp]\normalsize
  \centering
  \caption{Ablation Study on UAV123\cite{UAV123}. \textbf{Win} represents the multi-scale window transformer. \textbf{CS} denotes the proposed cyclic shifting strategy. \textbf{SR} means to apply the spatially regularized attention mask. \textbf{Pos} represents the relative position encoding.}
  \label{tab:3}
  \setlength{\tabcolsep}{2mm}{
    \begin{tabular}{c|cccc|cc}
    \toprule
    \#&Win&CS&SR&Pos&AUC&Prec.\\
    \midrule
    1&\multicolumn{4}{c|}{Original Transformer}&66.2&86.6\\
    \hline
    2&\checkmark&&&&54.4&70.8\\
    3&\checkmark&\checkmark&&&69.7&89.2\\
    4&\checkmark&\checkmark&\checkmark&&70.1&89.8\\
    5&\checkmark&\checkmark&&\checkmark&69.8&89.6\\
    6&\checkmark&\checkmark&\checkmark&\checkmark&70.5&90.3\\
    \bottomrule
    \end{tabular}
  }
\end{table}

\noindent \textbf{Effects of different components in our method.} We evaluate the effect of components including multi-scale window attention (Win), cyclic shifts (CS), spatially regularized attention mask (SR), and relative position encoding (Pos) employed in our method. The ablation study result is shown in Tab. \ref{tab:3}, \#1 represents the performance of the original transformer. We can see that window-level attention alone (\#2) is very ineffective as it greatly reduces the resolution of the attention mechanism, however, combining window-level attention with the cyclic shifting strategy can handle this drawback. It can be seen in \#3, there is a 15.3\% improvement in the AUC score after applying the cyclic shifts, and it outperforms the original transformer by 3.5\%, which illustrates that the cyclic shifting strategy plays a key role on the window-level attention. \#4 shows that the AUC score can be improved by 0.4\% when employing the spatially regularized mask to cyclic shifting samples, which demonstrates that spatial regularity can alleviate the boundary artifacts to a certain extent and improve the performance of window attention. In addition, we test the effectiveness of relative position encoding in our method following the way of \cite{SwinTransformer}. The performance improves by 0.1\% when the relative position encoding (\#5) is used, the small improvement indicates that the position encoding in window-level attention is not very important, and confirms that window-level attention itself contains rich position information.

\begin{table}[htbp]\normalsize
  \centering
  \caption{Comparison between the different window sizes on UAV123\cite{UAV123}. The first four items are adopting the same window size in each head. Multi-scale denotes the proposed CSWinTT.}
  \label{tab:4}
  \setlength{\tabcolsep}{4mm}{
    \begin{tabular}{c|cc}
    \toprule
    Window Size&AUC&Prec.\\
    \midrule
    $1\times 1$&66.2&86.6\\
    $2\times 2$&68.3&87.9\\
    $4\times 4$&70.0&89.6\\
    $8\times 8$&69.3&89.2\\
    Multi-scale&70.5&90.3\\
    \bottomrule
    \end{tabular}
  }
\end{table}

\noindent \textbf{Different window sizes for our transformer.}
To explore the performance of diverse window sizes on cyclic shifting window attention, we designed a quantitative analysis experiment as shown in Table \ref{tab:4}. The first four rows indicate that the same window size is used in all 8 heads in the case that cyclic shifting strategy is employed. It can see from the experimental results that the highest 70.0\% AUC score is obtained in size $4\times 4$ when using a single window size, as a matter of fact, the performance is really closer for all windows sizes. When adopting the multi-scale window size, the best AUC score of 70.5 is achieved, demonstrating that multi-scale windows can fuse information from different scales to improve the performance of the tracker.

\begin{table}[htbp]\normalsize
  \centering
  \caption{Comparison about the tracking speed for three computation optimization. \textbf{RMQ} represents removing the cyclic shifts of Query. \textbf{Peri} denotes halving shifting periods. \textbf{Prog} means adopting the programming optimization for matrix translation.}
  \label{tab:5}
  \setlength{\tabcolsep}{2mm}{
    \begin{tabular}{c|ccc|c}
    \toprule
    \#&RMQ&Peri&Prog&Speed(FPS)\\
    \midrule
    1&&&&1.0\\
    2&\checkmark&&&8.2\\
    3&\checkmark&\checkmark&&10.9\\
    4&\checkmark&\checkmark&\checkmark&12.4\\
    \hline
    5&\multicolumn{3}{c|}{Original Transformer}&14.9\\
    \bottomrule
    \end{tabular}
  }
\end{table}

\noindent \textbf{Computation optimization and speed analysis.}
Cyclic shifting strategy brings a large computational burden, we improve the tracking speed by applying some optimization strategies, including removing the cyclic shifts of Query (RMQ), halving the shifts periods (Peri), and adopting the programming optimization for matrix translation (Prog), Table \ref{tab:5} shows the effect of each optimization method. The tracking speed is around 1 FPS with no optimization adopted, as shown in \#1, which is almost an unusable state. With the cyclic shifts of Query are removed (\#2), the tracking speed is greatly improved to 8.2 FPS, and it can be further improved by halving the shifts periods (\#3). In addition, we also apply a PyTorch programming trick to use permutations of matrix coordinates to perform cyclic shifts instead of direct translations on the matrix, which also improves the tracking speed to some extent as shown in \#4. Due to the absolute amount of computation introduced by the cyclic shifting window attention, the computing efficiency of our method is not as good as the original transformer (\#5), but a satisfactory tracking speed of 12.4 FPS is achieved after our computational optimization.

\subsection{Qualitative Analysis}

\begin{figure}
  \centering
  \begin{subfigure}{0.325\linewidth}
    \includegraphics[width=\linewidth,keepaspectratio,page=6]{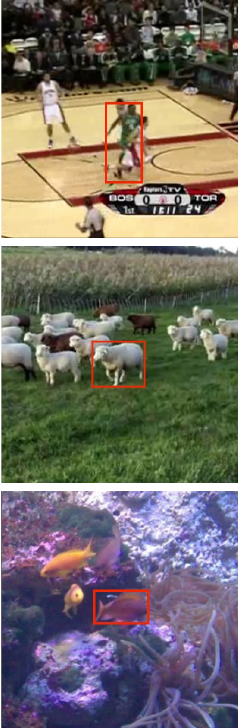}
    \subcaption*{Search Region}
    \label{fig:5a}
  \end{subfigure}
  \hfill
  \begin{subfigure}{0.325\linewidth}
    \includegraphics[width=\linewidth,keepaspectratio,page=7]{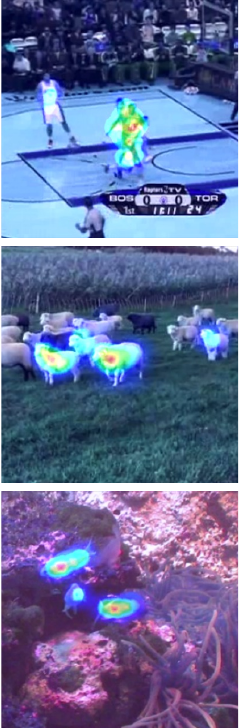}
    \subcaption*{Original Transformer}
    \label{fig:5b}
  \end{subfigure}
  \hfill
  \begin{subfigure}{0.325\linewidth}
    \includegraphics[width=\linewidth,keepaspectratio,page=8]{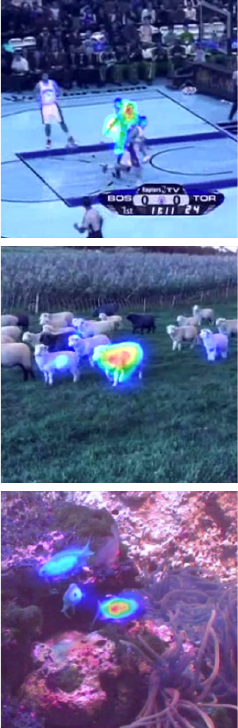}
    \subcaption*{Our Transformer}
    \label{fig:5c}
  \end{subfigure}
  \hfill
  \caption{Visual heat maps of the attention obtained by the original transformer (middle), and our proposed transformer (right). The red box indicates the target object in the search region (left).}
  \label{fig:5}
\end{figure}

Figure \ref{fig:5} shows the visual heat map of the attention, which exhibits the attention score of the last layer in the transformer matching module. The red area in the heat map indicates a high attention degree, while the blue area indicates a low attention degree. The first row shows the situation where the target object is obscured, the second and third rows show the scenario where the target is surrounded by similar distractors. From the visualization we can see that, compared to the pixel-level attention, the cyclic shifting window attention has a stronger discrimination ability of visual tracking, especially when the occlusion occurs or when there are similar distractors around the target object.

We further discuss why our proposed CSWinTT works. The strong discriminative ability mainly comes from two strategies: multi-scale window partition and cyclic shifts. After window partition, the target is split into multiple small blocks and each block contains the indivisible information of the object part. These blocks do not disrupt the pixels inside during attention, when some blocks are obscured and not visible, another part of the block can do attention without interference. Although there is no information exchange between different windows, the fusion of multi-scale windows can alleviate the problem, as well as be more robust to diverse sizes of occlusion areas. Additionally, the cyclic shifts can generate a more accurate attention score. For example, after window partition for a human body, there are two windows needed to do the attention. Suppose the first one is a window in the template that contains a head of the human body, which is in the center of the window. The second one is in the search region that contains the same head, as the human movement through the sequence, the head translates from the center to the edge of the window. At this point, a lower matching score will be obtained by window-level attention, which does not fully utilize the information in the windows. After employing the cyclic shifts, as shown in Figure \ref{fig:cs}, the head at the center of the template window and the head at the edge of the search region window can be finely matched. In addition, the position information in the attention can be obtained by the shift size, and this window-level position can better assist the tracking algorithm to distinguish the target object from the distractors.

\section{Conclusion}
In this work, we propose a transformer tracker with multi-scale cyclic shifting window attention, which is able to keep the integrity of the object and retain more location information when calculating the cross-window attention between the tracking target and the search area. Moreover, this new window attention is deliberated designed with two improvement schemes including spatially regularized attention mask and redundant computation removal to fully exploit the transformer structure for object tracking. Numerous experimental results on five challenging benchmarks demonstrate that our tracker performs better than previous state-of-the-art trackers. The proposed cyclic shifting window attention has stronger discrimination than the original pixel-level attention in the tracking field. Many other applications like image recognition and stereo matching may benefit from this window attention too. 

\section*{Acknowledgement}
This work is supported by the national key research and development program of China under Grant No.2020YFB1805601. 


{\small
\bibliographystyle{ieee_fullname}
\bibliography{egbib}
}

\end{document}